\documentclass[conference]{IEEEtran}
\usepackage{cite}
\usepackage{amsmath,amssymb,amsfonts}
\usepackage{algorithmic}
\usepackage{graphicx}
\usepackage{textcomp}
\usepackage{xcolor}
\def\BibTeX{{\rm B\kern-.05em{\sc i\kern-.025em b}\kern-.08em
    T\kern-.1667em\lower.7ex\hbox{E}\kern-.125emX}}

\usepackage{hyperref}
\usepackage{gensymb}
\usepackage{pifont}
\newcommand{\cmark}{\ding{51}}
\newcommand{\xmark}{\ding{55}}
\usepackage{pgfplots}
\pgfplotsset{compat=1.17}

\ifCLASSOPTIONcompsoc
\usepackage[caption=false,font=normalsize,labelfont=sf,textfont=sf]{subfig}
\else
\usepackage[caption=false,font=footnotesize]{subfig}
\fi

\begin{document}

\title{GazeSCRNN: Event-based Near-eye Gaze Tracking using a Spiking Neural Network}

\author{
\IEEEauthorblockN{Stijn Groenen}
\IEEEauthorblockA{\textit{Radboud University} \\
Nijmegen, The Netherlands \\
ORCID: 0009-0003-7042-2954}
\and
\IEEEauthorblockN{Marzieh Hassanshahi Varposhti}
\IEEEauthorblockA{\textit{Donders Institute for Brain,} \\
\textit{Cognition, and Behaviour} \\
Nijmegen, The Netherlands\\
ORCID: 0009-0005-7958-2051}
\and
\IEEEauthorblockN{Mahyar Shahsavari}
\IEEEauthorblockA{\textit{Donders Institute for Brain,} \\
\textit{Cognition, and Behaviour} \\
Nijmegen, The Netherlands\\
ORCID: 0000-0001-7703-6835}
}

\maketitle

\begin{abstract}
This work introduces GazeSCRNN, a novel spiking convolutional recurrent neural network designed for event-based near-eye gaze tracking. Leveraging the high temporal resolution, energy efficiency, and compatibility of Dynamic Vision Sensor (DVS) cameras with event-based systems, GazeSCRNN uses a spiking neural network (SNN) to address the limitations of traditional gaze-tracking systems in capturing dynamic movements. The proposed model processes event streams from DVS cameras using Adaptive Leaky-Integrate-and-Fire (ALIF) neurons and a hybrid architecture optimized for spatio-temporal data. Extensive evaluations on the EV-Eye dataset demonstrate the model's accuracy in predicting gaze vectors. In addition, we conducted ablation studies to reveal the importance of the ALIF neurons, dynamic event framing, and training techniques, such as Forward-Propagation-Through-Time, in enhancing overall system performance.
The most accurate model achieved a Mean Angle Error (MAE) of 6.034$\degree$ and a Mean Pupil Error (MPE) of 2.094 mm.
Consequently, this work is pioneering in demonstrating the feasibility of using SNNs for event-based gaze tracking, while shedding light on critical challenges and opportunities for further improvement.
\end{abstract}


\section{Introduction}
Eye and gaze tracking play a crucial role in various fields, including human-computer interaction, robotics, accessibility for users with motor disabilities, virtual reality (VR), and mixed reality (MR). These technologies leverage eye movements to create more intuitive and immersive experiences. However, traditional eye-tracking systems often face limitations in speed and energy efficiency~\cite{hansenEyeBeholderSurvey2010,liuEyeBeholderSurvey2022,duchowskiEyeTrackingMethodology2017}.

To address these limitations, Dynamic Vision Sensor (DVS) cameras have emerged as a promising alternative. These event-based cameras offer advantages over conventional frame-based cameras, such as high temporal resolution, low latency, and energy efficiency. DVS cameras achieve this by registering and emitting events only when changes in pixel brightness occur, unlike traditional cameras that capture images at a fixed frame rate regardless of scene changes~\cite{gallegoEventBasedVisionSurvey2022,lichtsteiner128x128120DB2008}.

Spiking Neural Networks (SNNs) are uniquely suited to process the event-driven data generated by DVS cameras. SNNs operate using sparse computations, where information is encoded and transmitted as discrete spikes, mirroring the biological neural processes in the human brain \cite{shahsavariAdvancementsSpikingNeural2023}. This results in significant gains in processing speed and energy efficiency, making SNNs well-suited for handling the spatio-temporal data from DVS cameras~\cite{pfeifferDeepLearningSpiking2018}.

This work presents a pioneering investigation into the potential of DVS cameras and SNNs for fast and efficient near-eye gaze tracking. The research focuses on a distinct SNN architecture: a spiking convolutional recurrent neural network, which we designed to predict the user's gaze vector.

The study uses the EV-Eye dataset~\cite{zhaoEVEyeRethinkingHighfrequency2023}, which comprises event streams recorded by DVS cameras aimed at participants' eyes during several gaze-related tasks.
During data preprocessing, the events are aggregated into frames and aligned with the ground truth gaze references.
Model performance is evaluated based on two primary metrics: Mean Pupil Error (MPE), which measures the distance between the predicted origin point and the actual pupil coordinates, and Mean Angle Error (MAE), which quantifies the angle between the predicted and ground truth gaze vectors.

The training process utilizes backpropagation and stochastic gradient descent, incorporating the surrogate gradient method to manage the non-differentiable nature of spikes~\cite{neftciSurrogateGradientLearning2019}. Given the time-dependent characteristics of the models, either truncated Backpropagation-Through-Time (BPTT) or Forward-Propagation-Through-Time (FPTT) is employed, allowing for effective training while managing memory constraints~\cite{bellecLongShorttermMemory2018,yinAccurateOnlineTraining2022}.

The source code for our models and experiments is available at \url{https://github.com/StijnGroenen/GazeSCRNN}.

\begin{figure*}[t]
\centering
\subfloat[Events]{\includegraphics[width=0.22\linewidth]{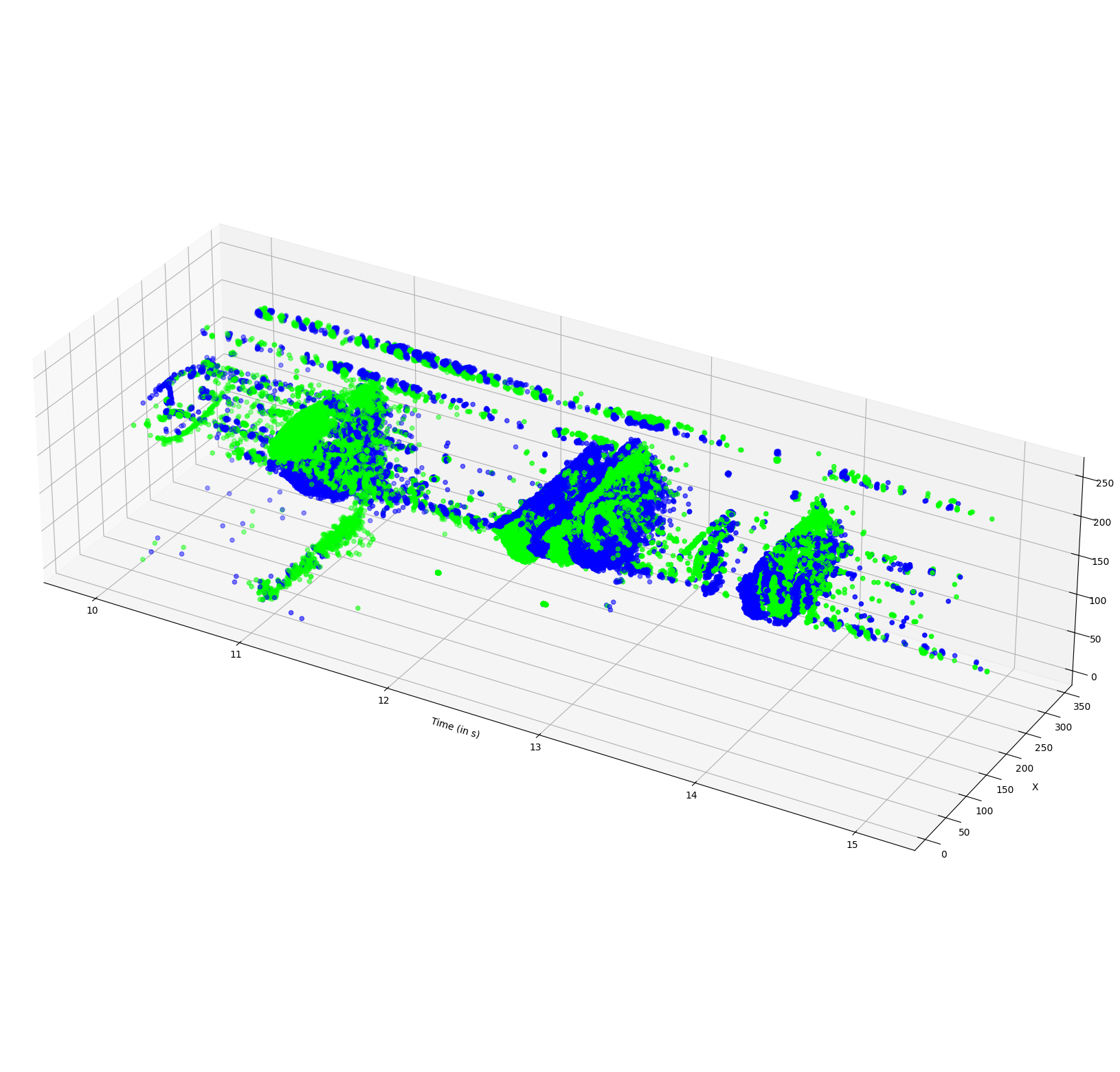}
\label{fig:events_plot}}
\hfil
\subfloat[Frames]{\includegraphics[width=0.22\linewidth]{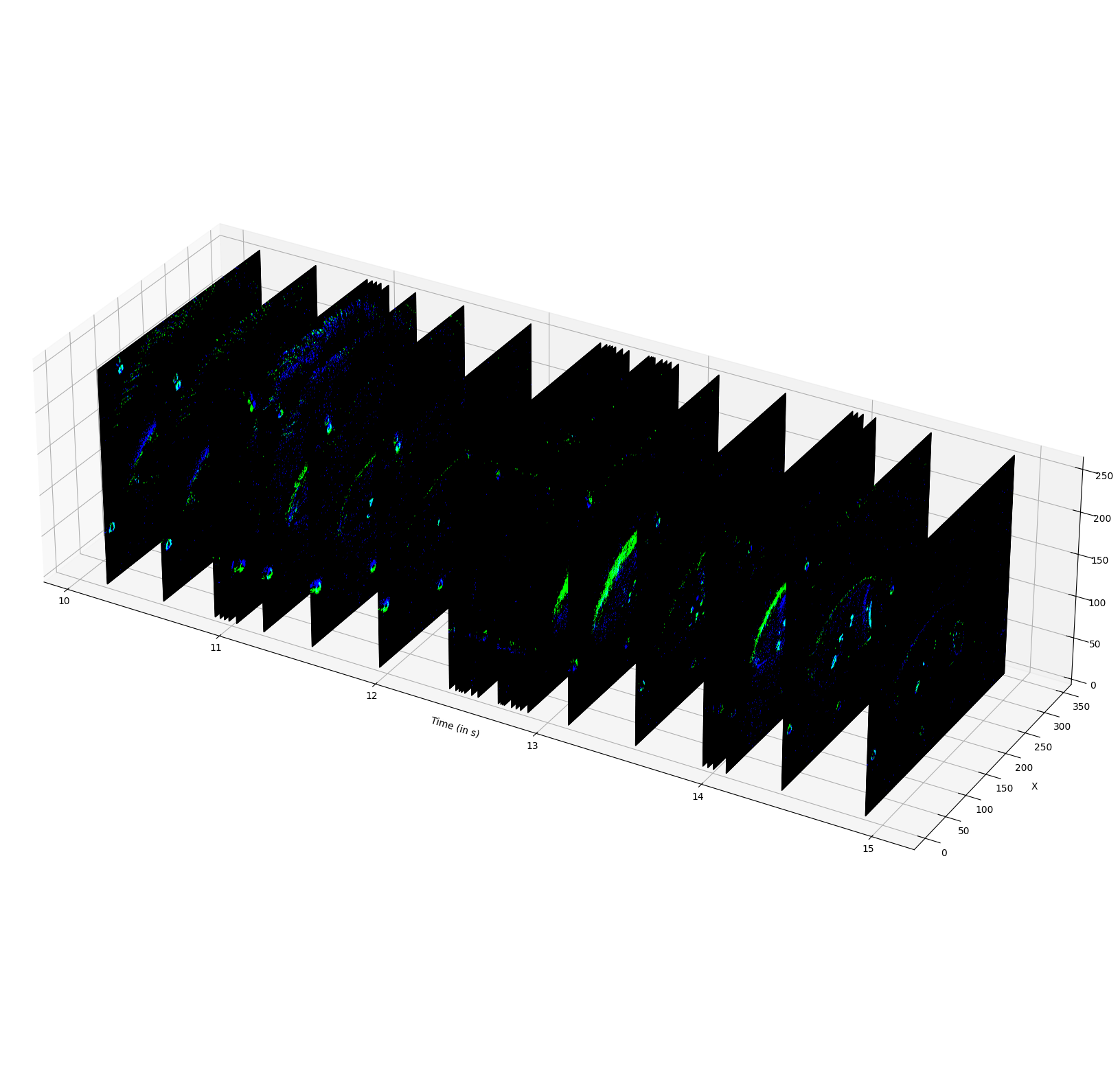}
\label{fig:frames_plot}}
\hfil
\subfloat[Spike pattern]{\includegraphics[width=0.22\linewidth]{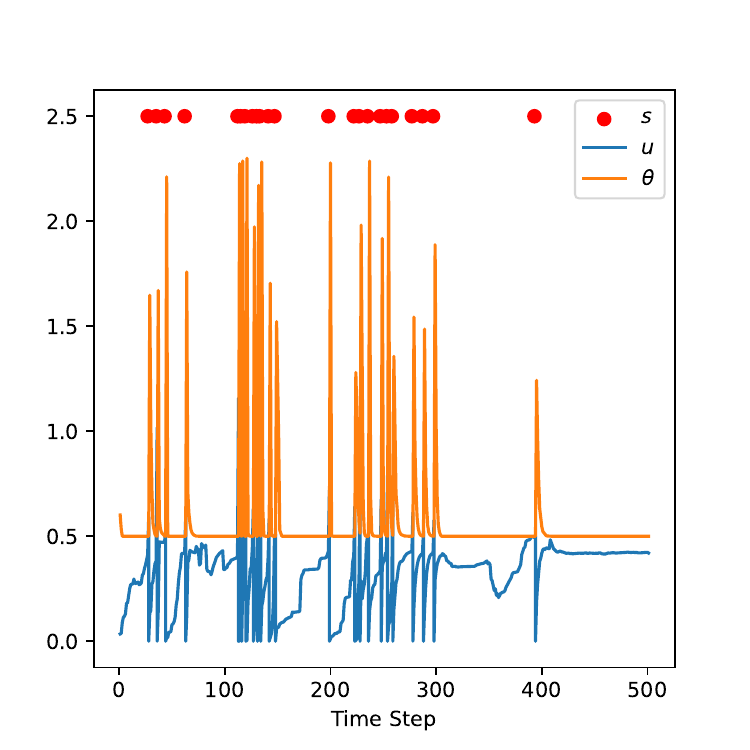}
\label{fig:spike_plot}}
\hfil
\subfloat[Gaze vectors]{\includegraphics[width=0.22\linewidth]{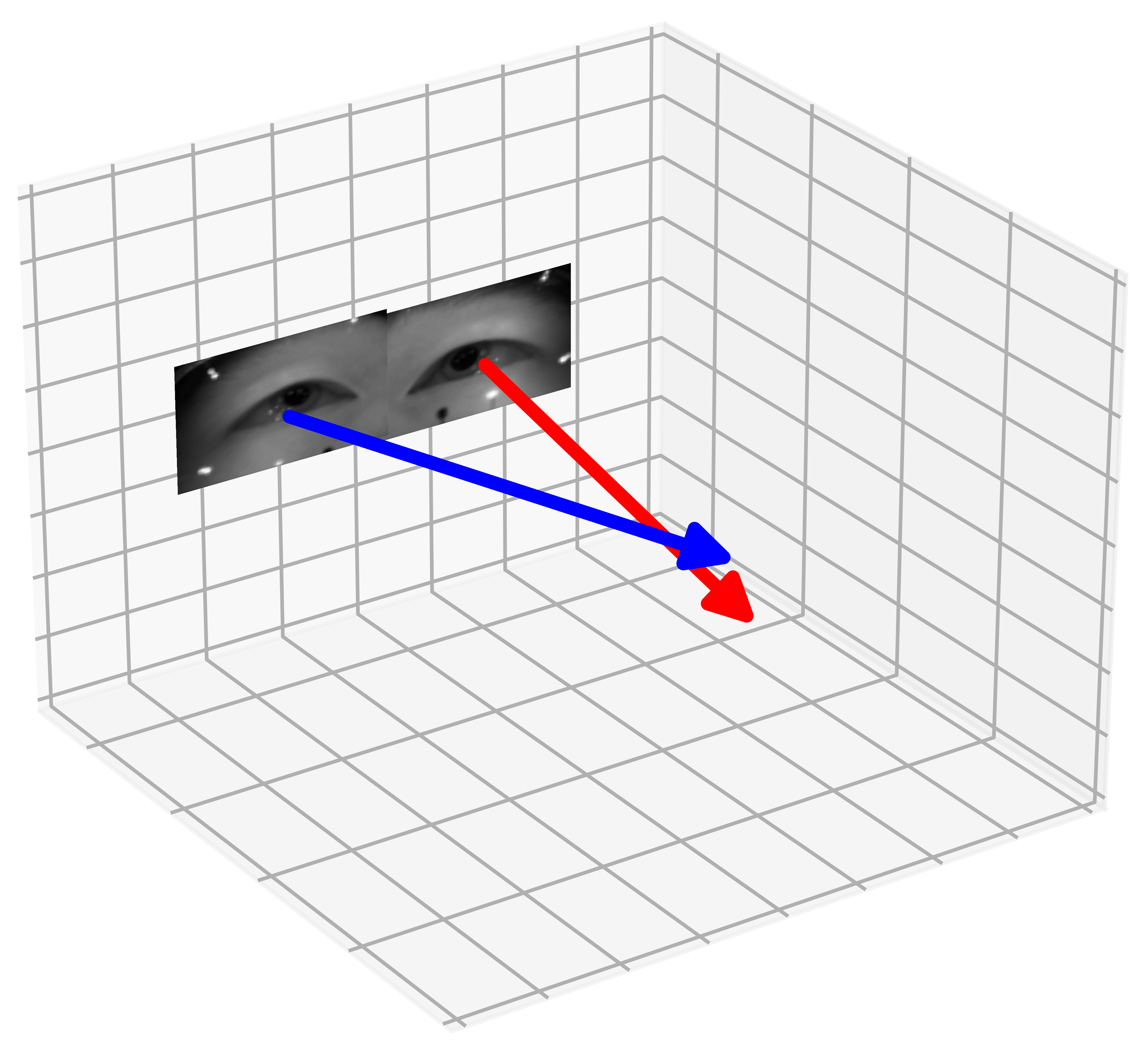}
\label{fig:gaze_vectors_plot}}
\caption{
Overview of the event-based gaze tracking pipeline.
\ref{fig:events_plot} shows input events captured by a Dynamic Vision Sensor (DVS) camera.
\ref{fig:frames_plot} shows the events aggregated into frames during data preprocessing.
\ref{fig:spike_plot} shows frames processed by the spiking neural network, producing sequences of complex spike patterns.
\ref{fig:gaze_vectors_plot} shows output gaze vectors, indicating the user's point of focus.
}
\label{fig:event_based_gaze_tracking_pipeline}
\end{figure*}

\section{Related work}
\label{section:related_works}

\subsection{Gaze tracking}
Gaze-tracking technologies have been categorized by Liu et al.~\cite{liuEyeBeholderSurvey2022} into two primary approaches: model-based and appearance-based techniques.
These approaches address the gaze tracking task through fundamentally different strategies, each with unique benefits and limitations based on the application environment, hardware requirements, and tolerance for head motion. Additionally, gaze tracking systems can be divided into remote setups, which rely on an external camera, and near-eye or head-mounted setups, for which the camera moves with the user's head, limiting the increased difficulty of handling head movements~\cite{liuEyeBeholderSurvey2022}.

Model-based gaze tracking methods use geometric models of the eye to estimate gaze direction, relying on specific eye features such as the pupil and iris location, eye corner points, and glints from (infrared) illumination.
Model-based methods can be categorized into 2D mapping-based and 3D model-based approaches~\cite{liuEyeBeholderSurvey2022}.
For 2D mapping-based methods, a mapping function is constructed between certain 2D eye features, such as pupil or iris position, and gaze points on a screen.
This requires personal calibration, making it sensitive to head motion effects, since the mapping is based on a fixed head position relative to the screen.
Alternatively, 3D model-based methods leverage geometric relationships and the 3D structure of the eyeball to reconstruct the line-of-sight.

In contrast to model-based methods, appearance-based gaze tracking techniques rely on direct image analysis rather than explicit geometric models of the eye. These methods use machine learning for predicting the gaze direction based on the input image. This enables a more flexible approach suitable for a wide range of environments, often without a strict requirement for personal calibration or infrared illumination~\cite{jiangAppearanceBasedGazeTracking2019}.
Neural networks, particularly convolutional neural networks (CNNs), have shown promising results in appearance-based gaze tracking. CNNs use convolutional layers to automatically extract hierarchical features from eye images, making them robust to variations in lighting and head poses.
However, they require extensive computational resources and a large labelled dataset for effective training~\cite{jiangAppearanceBasedGazeTracking2019}.
Furthermore, challenges remain in achieving real-time performance on lightweight devices.
In this paper, we train a spiking neural network (SNN) for event-based near-eye gaze tracking.
Our solution does not use explicit user calibration. However, it is possible to apply personal calibration through an online learning approach, allowing the SNN to become more optimized over time for a specific user.

\subsection{Event-based Vision and Spiking Neural Networks for Gaze Tracking}
Dynamic Vision Sensor (DVS) cameras and Spiking Neural Networks (SNNs) represent a promising combination. DVS cameras emit events only when changes in pixel brightness occur, which provides high temporal resolution, low latency, and energy efficiency~\cite{gallegoEventBasedVisionSurvey2022,lichtsteiner128x128120DB2008}. These characteristics align well with the sparse and event-driven processing model of SNNs, which encode information as discrete spikes that mimic biological neural activity. This can lead to significant improvements in computational costs and energy efficiency compared to traditional artificial neural networks, especially when implemented on neuromorphic hardware~\cite{pfeifferDeepLearningSpiking2018}. 

Despite this natural synergy, much of the prior work with DVS cameras for eye and gaze tracking has relied on conventional processing techniques. For example, Angelopoulos et al.~\cite{angelopoulosEventBasedNearEyeGaze2021} combined regular frames and DVS events to dynamically update a pupil model at high speeds, and then estimate the corresponding gaze vectors using a user-calibrated polynomial regression. Ryan et al.~\cite{ryanRealtimeFaceEye2021} reconstructed frames from DVS events, and applied a convolutional recurrent neural network to predict face and eye positions. Similarly, Chen et al.~\cite{chen3ETEfficientEventbased2023} proposed a Change-Based ConvLSTM network for efficient pupil tracking.

Several works highlight the potential of integrating SNNs and DVS cameras for eye tracking systems. Jiang et al.~\cite{jiangEyeTrackingBased2024a} demonstrated an SNN-based near-eye eye-tracking model that leverages the asynchronous temporal information from event cameras, achieving enhanced stability and accuracy with low computational costs. Similarly, Bonazzi et al.~\cite{bonazziRetinaLowPowerEye2024} proposed an SNN for pupil tracking, implemented on neuromorphic hardware, to exploit the low latency and energy efficiency of the event-based paradigm. However, prior research into SNNs was limited to eye tracking, rather than gaze tracking.

This study builds upon all these efforts by proposing a novel spiking convolutional recurrent neural network architecture optimized for event-based near-eye gaze tracking, leveraging the temporal precision and energy efficiency of DVS cameras, as well as the sparse, and energy-efficient computations of SNNs, with the capability to be implemented on neuromorphic and edge devices for real-time gaze tracking and online learning.

\section{Methods}
\label{section:methods}

\subsection{EV-Eye Dataset}
\label{section:dataset}
For training and evaluating our model, we use the event-based EV-Eye dataset introduced by Zhao et al. (2023)~\cite{zhaoEVEyeRethinkingHighfrequency2023}.
The dataset features asynchronous event streams collected using a pair of DAVIS346 Dynamic Vision Sensor (DVS) cameras with a resolution of $346 \times 260$ \cite{inivationagDAVIS346Documentation2024} (see Figure~\ref{fig:events_plot}), one dedicated to each eye, positioned to capture participants' eye movements while they perform specific gaze tasks.

The EV-Eye dataset covers three gaze tasks that simulate common eye movement patterns~\cite{zhaoEVEyeRethinkingHighfrequency2023}:
\begin{enumerate}
    \item Smooth Pursuit: Participants follow a moving stimulus across a screen in a smooth, predictable trajectory. 
    \item Random Saccade: Participants search for a stimulus that appears at random locations on the screen.
    \item Fixation: Participants maintain their gaze on a stationary stimulus for an extended period, achieved by keeping the stimulus in one position during the random saccade task.
\end{enumerate}

Data was gathered from 48 participants, each completing two sessions for both the smooth pursuit and random saccade tasks, resulting in a total of 192 event streams. Alongside the event data, the dataset provides gaze references in the form of vectors recorded by Tobii Pro Glasses 3 at 100Hz \cite{tobiiabTobiiProGlasses2024}, as depicted in Figure~\ref{fig:gaze_vectors_plot}. Zhao et al.~\cite{zhaoEVEyeRethinkingHighfrequency2023} argue that gaze vectors provide a more accurate reference compared to screen coordinates, noting that eye movements often diverge from the exact location of the stimulus on a screen, particularly during the search phase of the random saccade task.

\subsubsection{Data Preprocessing}
\label{section:preprocessing}
Prior to training our model, we preprocess the event streams from the DVS cameras by aggregating them into frames (see Figure~\ref{fig:frames_plot}). This aggregation is either performed using fixed time windows or by using a fixed number of events per frame (see Section~\ref{section:experiment_framing_method}).

For most of our models, the frames are downscaled by a factor of two in order to improve efficiency, resulting in final dimensions of $173 \times 130 \times 2$. The two channels represent the positive and negative polarity events, respectively.

The gaze references provided in the EV-Eye dataset are utilized to compute the loss function. These references consist of a 3D origin point and two spherical coordinates representing the gaze direction (omitting the radius). The spherical coordinates are normalized so that $0\degree$ corresponds to the user looking straight ahead. The gaze references are aligned with the aggregated frames as closely as possible using their timestamps, and are linearly interpolated between true references when the timestamps do not line up exactly.

However, since the gaze references in the EV-Eye dataset are predictions from another gaze tracker, they may not perfectly reflect the user's true gaze. Furthermore, we observed prolonged periods where the device failed to provide valid gaze vectors (max. $8.5\text{s}$), leading to interpolation between widely separated ground truth values. These interpolated values can be unreliable. To address this, we optionally mask these interpolated references for some models, if the time window between the ground truth references exceeds a pre-defined threshold, excluding them from loss and metric computations (see Section~\ref{section:experiment_unreliable_references}).  

To manage memory constraints and enable fixed-size training batches, we divide the event streams into sequences of 1000 frames. These sequences are organized into training ($70\%$), validation ($15\%$), and testing ($15\%$) sets by randomly selecting sequences for validation and testing using a fixed seed for reproducibility.

\subsection{Spiking Neurons}
\label{section:spiking_neurons}

\begin{figure}[t]
\centerline{\includegraphics[width=\linewidth]{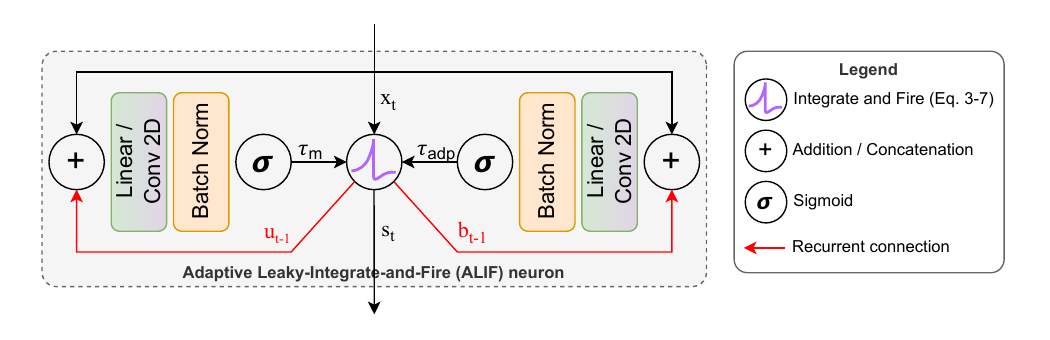}}
\caption{Adaptive Leaky-Integrate-and-Fire Neuron with Liquid Time Constants}
\label{fig:spiking_neuron}
\end{figure}

There are various types of spiking neurons.
The most commonly used spiking neuron model is the Leaky-Integrate-and-Fire (LIF) neuron. As the name suggests, LIF-neurons integrate the incoming spikes to update a leaky membrane potential which decays over time, and it fires (and resets) when the membrane potential reaches a threshold.
Instead of regular Leaky-Integrate-and-Fire (LIF) neurons, our models use two distinct types of spiking neuron models: Parametric Leaky-Integrate-and-Fire (PLIF) neurons and Adaptive Leaky-Integrate-and-Fire (ALIF) neurons.
ALIF neurons are adaptive, meaning their spiking behaviour responds dynamically to the input (see Figure~\ref{fig:spike_plot}). We base our implementation on the models by Yin et al.~\cite{yinEffectiveEfficientComputation2020,yinAccurateOnlineTraining2022}, which incorporate Liquid Time Constants (LTCs) that enable dynamic adjustment of both membrane potential decay and the firing threshold.

In these adaptive neurons, the membrane potential decay constant $\tau_{m}$ and the adaptation decay constant $\tau_{adp}$ are computed at each time step based on the input and neuron state. This gives ALIF neurons liquid time constants that vary with the network's input, which allows them to track both short- and longer-term dependencies within the sequence. The result is a neuron model that captures richer temporal dynamics than standard spiking neurons~\cite{yinEffectiveEfficientComputation2020}.

PLIF neurons, in contrast, use fixed parameters for $\tau_{m}$ and $\tau_{adp}$, which are optimized during training but do not vary with input at each step.

Equations~\ref{eq:tau_m} through~\ref{eq:reset} describe the computations governing both types of neurons.

\begin{equation}
    \tau_{m} = 
    \begin{cases}
        \sigma(\text{Conv2D}(x_{t} + u_{t-1})) & \text{if} \ x_{t} \ \text{is 2D} \\
        \sigma(\text{Linear}(x_{t} \mathbin\| u_{t-1})) & \text{if} \ x_{t} \ \text{is 1D} \\
    \end{cases}
    \label{eq:tau_m}
\end{equation}
\begin{equation}
    \tau_{adp} = 
    \begin{cases}
        \sigma(\text{Conv2D}(x_{t} + b_{t-1})) & \text{if} \ x_{t} \ \text{is 2D} \\
        \sigma(\text{Linear}(x_{t} \mathbin\| b_{t-1})) & \text{if} \ x_{t} \ \text{is 1D} \\
    \end{cases}
    \label{eq:tau_adp}
\end{equation}
\begin{equation}
    u_{t} = u_{t-1} + \tau_{m}(x_{t} - u_{t-1})
    \label{eq:membrane}
\end{equation}
\begin{equation}
    b_{t} = \tau_{adp} b_{t-1} + (1 - \tau_{adp}) s_{t-1}
    \label{eq:b_t}
\end{equation}
\begin{equation}
    \theta_{t} = b_{0} + \beta b_{t}
    \label{eq:threshold}
\end{equation}
\begin{equation}
    s_{t} = \Theta(u_{t} - \theta_{t})
    \label{eq:spike}
\end{equation}
\begin{equation}
    u_{t} = u_{t}(1 - s_{t}) + u_{r}s_{t}
    \label{eq:reset}
\end{equation}

The parameter $\tau_{m}$ in Equation~\ref{eq:tau_m} is the membrane time constant for the neuron, controlling the rate at which the neuron's membrane potential $u_t$ decays over time. For ALIF neurons, the time constant can vary adaptively at each time step and is calculated as a function of the current input $x_t$ and the previous membrane potential $u_{t-1}$. Specifically, if the input $x_t$ is two-dimensional (excluding channels; such as spatially structured data, like frames), a sigmoid function $\sigma$ and convolutional layer are used to compute the time constants, whereas for one-dimensional inputs, a fully-connected layer is used. For PLIF neurons, $\tau_{m}$ is not adaptive, but is a trained parameter.

The parameter $\tau_{adp}$ in Equation~\ref{eq:tau_adp} is the adaptation time constant, which controls the rate at which the adaptation variable $b_t$ decays. Like $\tau_{m}$, for ALIF neurons, this time constant is also input-dependent, computed based on $x_t$ and the adaptation variable from the previous time step $b_{t-1}$.

The parameter $u_t$ in Equation~\ref{eq:membrane} is the membrane potential of the neuron at time $t$, representing the internal state that integrates input signals. It is updated based on the previous membrane potential $u_{t-1}$, the current input $x_t$, and the membrane time constant $\tau_{m}$, which scales the influence of the current input on the potential decay rate.

The variable $b_t$ in Equation~\ref{eq:b_t} is an adaptation variable that represents the dynamic adjustment of the neuron's firing threshold. It is updated based on the previous adaptation state $b_{t-1}$, the adaptation time constant $\tau_{adp}$, and the previous spike output $s_{t-1}$. This adaptation variable increases with spiking activity, raising the threshold temporarily to reduce the firing rate.

The firing threshold $\theta_t$ in Equation~\ref{eq:threshold} at time $t$ is defined as the baseline threshold $b_0$ plus a scaling factor $\beta$ applied to the adaptation variable $b_t$. This adaptive threshold adjusts dynamically to control the neuron's sensitivity to inputs, preventing excessive spiking. Like in~\cite{yinEffectiveEfficientComputation2020}, $\beta$ is a hyperparameter that defaults to $1.8$.

The binary variable $s_t$ in Equation~\ref{eq:spike} represents the spike output of the neuron at time $t$. It is computed by the Heaviside step function $\Theta$, which outputs 1 (a spike) if the membrane potential $u_t$ exceeds the threshold $\theta_t$, and 0 otherwise.

Equation~\ref{eq:reset} updates the membrane potential $u_t$ after a spike. When $s_t = 1$, indicating a spike, $u_t$ is reset to the resting potential $u_r$ (a hyperparameter that defaults to $0$), otherwise, $u_t$ remains unchanged. This mechanism ensures that the neuron's membrane potential is reset after each spike. In our implementation, it is also possible to disable the reset mechanism, for example for the final layer, where the membrane potential of the neurons can be used as the output of the model.

Figure~\ref{fig:spiking_neuron} depicts the internal structure of an ALIF neuron layer with Liquid Time Constants.

\subsection{GazeSCRNN: Spiking Convolutional Recurrent Neural Network Architecture}
\label{section:scrnn}

\begin{figure*}[t]
\centerline{\includegraphics[width=\linewidth]{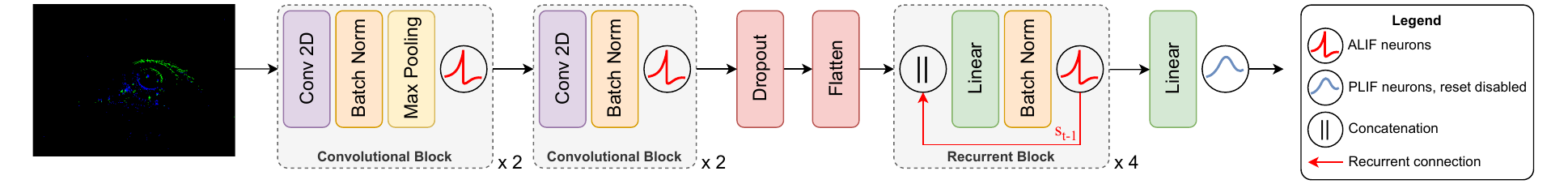}}
\caption{GazeSCRNN, a Spiking Convolutional Recurrent Neural Network architecture for event-based gaze tracking}
\label{fig:scrnn}
\end{figure*}

In this work, we propose the GazeSCRNN architecture: a spiking convolutional recurrent neural network designed for event-based gaze tracking. Since the event data contains both spatial and temporal dimensions, this architecture employs a combination of convolutional and recurrent layers to process the event data and predict the user's gaze vector. Adaptive Leaky-Integrate-and-Fire (ALIF) neurons, as described in Section~\ref{section:spiking_neurons}, are used in this model to allow the spiking behaviour of neurons to adapt to the input data and be optimized during training \cite{yinEffectiveEfficientComputation2020}, adding complexity to the otherwise relatively simple architecture.

Figure~\ref{fig:scrnn} shows a diagram of the GazeSCRNN architecture. The network accepts a 2-channel input tensor of dimensions $2 \times 130 \times 173$, representing the downsized aggregated frames derived from the events captured by the DVS camera. Spatial features are extracted through a series of four convolutional blocks, each comprising a convolutional layer, batch normalization, max-pooling (for the first two blocks), and ALIF neurons.

The convolutional blocks progressively reduce the spatial dimensions while increasing the number of filters to extract progressively richer features, starting with 32 filters in the first block and scaling up to 128 filters in the final block. Batch normalization is applied after each convolutional layer to enhance training stability, and max-pooling layers reduce the spatial dimensions for efficient processing. ALIF neuron layers integrate incoming values from the convolutional layers into their membrane potentials and fire when the spike threshold is exceeded, introducing sparsity and efficiency to the computation. Dropout regularization is applied after the last convolutional block to reduce overfitting.

Temporal dependencies in the event stream are modelled by four recurrent blocks, each consisting of a fully connected layer, batch normalization, and ALIF neurons. These recurrent blocks integrate temporal information by concatenating their input features with the spiking outputs from the previous time step, enabling the model to retain and refine temporal representations across multiple time steps.

The final gaze vector prediction is computed by a fully connected layer, which maps the processed features to five outputs, corresponding to the three origin coordinates and two angular components of the gaze vector. This layer is followed by a layer of Parametric Leaky-Integrate-and-Fire (PLIF) neurons with their reset mechanism disabled, meaning they do not fire or reset their membrane potentials. Since gaze tracking is a regression problem, the membrane potentials of this final PLIF layer are directly interpreted as the output gaze vector.

The combination of convolutional and recurrent blocks in this architecture, containing a total of $2.3\text{M}$ parameters, enables the network to effectively leverage both the spatial structure and temporal dynamics inherent in the input event streams. This synergy allows GazeSCRNN to process event-based data effectively, making it particularly well-suited for gaze tracking applications.

\subsection{Metrics and Loss}
\label{section:metrics_loss}
For gaze tracking, we predict gaze vectors. These vectors consist of an origin point, which consists of three coordinates in mm, and two spherical coordinates, representing the direction where the gaze vector is pointing.
Therefore, to measure the accuracy of the predicted gaze vectors, we compute two important metrics.
We compute the distance between the predicted origin point and the ground-truth pupil coordinates from the target data, which we will call the pupil error. Furthermore, we compute the angle between the predicted gaze vector and the ground-truth gaze vector from the target data, which we will call the angle error.
The accuracy of each model can thus be summarized by the Mean Pupil Error (MPE) and the Mean Angle Error (MAE).

To optimize our model, we use a loss function that minimizes these two errors.
For this, we use a combination of a Mean Squared Error Loss for the pupil distance, and a Cosine Distance loss for the angles.
For our experiments, we have chosen to select our best models based on their Mean Angle Error, because this is the metric that currently needs the most improvement.

In addition to the MPE and MAE metrics, we also measure the Mean Firing Rate (MFR) of the spiking neurons, which represents the mean ratio of neurons that spike during each time step, giving an indication of the sparsity of the spikes in the network.
More sparsity, meaning fewer spikes, can enhance processing efficiency, particularly on neuromorphic hardware. However, it is important to acknowledge that sparsity is not currently a primary optimization goal. However, it is possible to also optimize for sparsity by including it as an additional term in the loss.

\subsection{Training}
\label{section:training}
All our models are trained using backpropagation and stochastic gradient descent. Since the spikes generated by our spiking neurons are discontinuous and therefore non-differentiable, we employ the surrogate gradient method to compute the gradients for these spiking neurons~\cite{neftciSurrogateGradientLearning2019}.

As described in Section~\ref{section:scrnn}, our model is time-dependent, meaning that computation at each time step depends on prior time steps. Training models with such temporal dependencies often relies on Backpropagation-Through-Time (BPTT), which unrolls the entire network across all time steps to propagate the loss backward through the sequence. However, BPTT incurs substantial memory costs that grow with the input sequence length, making it impractical for long sequences, as in our case.

To address these challenges, we use two methods to manage memory and computational complexity: truncated Backpropagation-Through-Time and Forward-Propagation-Through-Time (FPTT).
In truncated BPTT, the loss is propagated back only to the inputs from the last few time steps, rather than the entire sequence. By selecting an optimal truncation length, we can balance memory consumption and computational costs, with a high enough backpropagation depth for good accuracy.

In addition to truncated BPTT, we apply Forward-Propagation-Through-Time (FPTT)~\cite{yinAccurateOnlineTraining2022,kagTrainingRecurrentNeural2021}, which offers a significant advantage for training adaptive spiking neural networks with long temporal dependencies. FPTT minimizes the computation and memory requirements associated with BPTT by computing updates based on an instantaneous loss function. This loss is adjusted with a dynamic regularization term based on a running average of the model parameters, which stabilizes the gradients without accumulating error from previous steps, as is often seen with BPTT in recurrent neural networks. This approach removes the need to unroll the network across all time steps, allowing FPTT to train on long sequences with fixed memory requirements~\cite{yinAccurateOnlineTraining2022,kagTrainingRecurrentNeural2021}. Since we are dealing with a sequence-to-sequence problem, computing the instantaneous loss at every time step is trivial. Furthermore, this immediate, time-localized updating not only minimizes memory usage but also supports online training, which can be beneficial for real-time applications such as gaze tracking, for example, to allow for personal calibration by the end user.

Yin et al. in~\cite{yinAccurateOnlineTraining2022} demonstrated that FPTT achieves improved accuracy and memory efficiency when paired with Adaptive Leaky-Integrate-and-Fire (ALIF) neurons with Liquid Time Constants (LTCs). In our model, LTCs allow neurons to dynamically adjust their behaviour, such as membrane potential decay and threshold adaptation, based on the inputs and prior neuronal states. This adaptability enables FPTT to optimize the spiking dynamics of the neurons, making it highly effective in capturing both short- and long-term dependencies with minimal computational overhead~\cite{yinAccurateOnlineTraining2022}, which is advantageous for resource-constrained applications like wearable and edge devices~\cite{varposhtiEnergyEfficientSpikingRecurrent2024}.

In our experiments, we apply both FPTT and truncated BPTT, as well as a hybrid approach that combines FPTT with backpropagation over a limited number of time steps. In Section~\ref{section:experiment_fptt}, we present an analysis of the effects of different backpropagation depths, as well as the benefits of FPTT for training our models efficiently and accurately.

\section{Ablation studies}
\label{section:ablation_studies}
This section presents ablation studies that assess the impact of specific features and hyperparameters on model accuracy (Mean Angle Error, MAE; Mean Pupil Error, MPE) and efficiency (Mean Firing Rate, MFR).

We use the GazeSCRNN architecture described in Section~\ref{section:scrnn}. Each experiment modifies a particular feature or hyperparameter while training on the training dataset and evaluating performance per epoch on the validation dataset. The final evaluation uses a previously unseen test dataset, selecting the model with the best MAE on the validation set, as MAE is deemed the most significant metric for model usability and accuracy at this time.

\subsection{Experiment 1: Effect of FPTT and BPTT Time Steps}
\label{section:experiment_fptt}
\begin{figure}
\centering
\begin{tikzpicture}
    \begin{axis}[
        width=0.9\linewidth,
        height=0.6\linewidth,
        xlabel={Truncated BPTT Time Steps ($T$)},
        xlabel style={font=\small},
        ylabel={Mean Angle Error (in degrees)},
        ylabel style={yshift=-1ex, color=blue, font=\small},
        xmin=0, xmax=17,
        ymin=5, ymax=14,
        xtick={1, 2, 4, 8, 12, 16},
        legend style={font=\scriptsize, legend columns=2},
        grid=both,
        grid style={dotted},
        tick label style={font=\small},
        every axis plot/.append style={thick},
        legend cell align={left},
        axis y line*=left,
        axis x line*=bottom
    ]

    \addplot[
        mark=o,
        color=blue,
        solid,
    ] coordinates {
        (1, 12.20)
        (2, 11.89)
        (4, 9.045)
        (8, 8.572)
        (12, 9.231)
        (16, 9.150)
    };
    \label{plt:mae_fptt}
    \addlegendentry{FPTT};

    \addplot[
        mark=o,
        color=blue,
        dashed,
    ] coordinates {
        (1, 86.25)
        (2, 86.26)
        (4, 9.424)
        (8, 9.189)
        (12, 8.463)
        (16, 9.193)
    };
    \label{plt:mae_no_fptt}
    \addlegendentry{No FPTT};
    
    \addlegendimage{/pgfplots/refstyle=plt:mpe_fptt}\addlegendentry{FPTT};
    \addlegendimage{/pgfplots/refstyle=plt:mpe_no_fptt}\addlegendentry{No FPTT};
    
    \end{axis}

    \begin{axis}[
        width=0.9\linewidth,
        height=0.6\linewidth,
        xlabel={Truncated BPTT Time Steps ($T$)},
        xlabel style={font=\small},
        ylabel={Mean Pupil Error (in mm)},
        ylabel style={yshift=1ex, color=orange, font=\small},
        xmin=0, xmax=17,
        ymin=1.8, ymax=5.2,
        xtick={1, 2, 4, 8, 12, 16},
        axis y line*=right,
        axis x line=none,
        ytick style={color=orange},
        yticklabel style={color=orange},
        every axis plot/.append style={thick},
    ]

    \addplot[
        mark=square,
        color=orange,
        solid,
    ] coordinates {
        (1, 3.382)
        (2, 3.919)
        (4, 2.746)
        (8, 3.116)
        (12, 3.814)
        (16, 2.976)
    };
    \label{plt:mpe_fptt}

    \addplot[
        mark=square,
        color=orange,
        dashed,
    ] coordinates {
        (1, 4.811)
        (2, 5.016)
        (4, 3.480)
        (8, 2.938)
        (12, 3.064)
        (16, 3.770)
    };
    \label{plt:mpe_no_fptt}
    \end{axis}
\end{tikzpicture}
\caption{Mean Angle Error (MAE) and Mean Pupil Error (MPE) vs. the number of Truncated Backpropagation-Through-Time Time Steps ($T$) for models trained with and without Forward-Propagation-Through-Time (FPTT)}
\label{fig:mae_mpe_fptt_timesteps}
\end{figure}
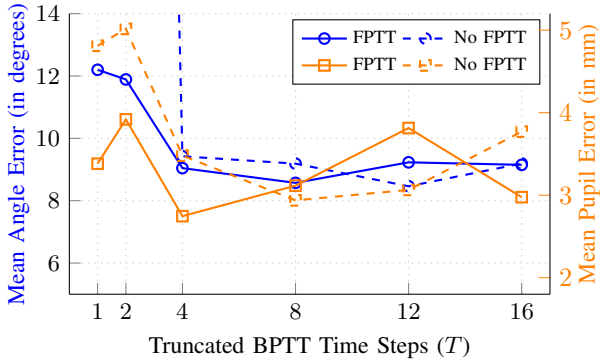

We investigate Forward-Propagation-Through-Time (FPTT) and variations in the number of truncated Backpropagation-Through-Time (BPTT) steps, denoted as $T$. In our implementation, training applies BPTT to the last $T$ time steps after waiting for $T$ steps during forward propagation. Increasing $T$ reduces the optimizer updates per epoch, as each input frame (representing one time step) propagates through the model only once. Results, shown in Table~\ref{tab:ablation_study_fptt_timesteps} and Figure~\ref{fig:mae_mpe_fptt_timesteps}, reveal that both FPTT and an optimal $T$ improve the accuracy of the model. Applying FPTT shows an improvement in accuracy, especially when $T$ is lower. As $T$ increases, the need for FPTT diminishes. For subsequent experiments, we will use FPTT with $T = 8$ to train the models.

\begin{table}[!t]
\caption{Ablation Study Results of Applying FPTT and Varying the Number of Truncated BPTT Time Steps for GazeSCRNN}
\label{tab:ablation_study_fptt_timesteps}
\centering
\scriptsize
\begin{tabular}{c|c||c|c|c}
\hline
\textbf{FPTT} & \textbf{BPTT Time Steps} & \textbf{MAE} & \textbf{MPE} & \textbf{MFR} \\
\hline
\hline
\cmark & $1$ & $12.20\degree$ & $3.382\text{mm}$ & $0.03246$ \\
\hline
\cmark & $2$ & $11.89\degree$ & $3.919\text{mm}$ & $0.03088$ \\
\hline
\cmark & $4$ & $9.045\degree$ & $2.746\text{mm}$ & $0.02437$ \\
\hline
\cmark & $8$ & $\mathbf{8.572\degree}$ & $3.116\text{mm}$ & $0.02244$ \\
\hline
\cmark & $12$ & $9.231\degree$ & $3.814\text{mm}$ & $0.03855$ \\
\hline
\cmark & $16$ & $9.150\degree$ & $2.976\text{mm}$ & $0.02237$ \\
\hline
\xmark & $1$ & $86.25\degree$ & $4.811\text{mm}$ & $0.03118$ \\
\hline
\xmark & $2$ & $86.26\degree$ & $5.016\text{mm}$ & $0.02432$ \\
\hline
\xmark & $4$ & $9.424\degree$ & $3.480\text{mm}$ & $0.02416$ \\
\hline
\xmark & $8$ & $9.189\degree$ & $2.938\text{mm}$ & $0.03492$ \\
\hline
\xmark & $12$ & $\mathbf{8.463\degree}$ & $3.064\text{mm}$ & $0.03457$ \\
\hline
\xmark & $16$ & $9.193\degree$ & $3.770\text{mm}$ & $0.03592$ \\
\hline
\end{tabular}
\end{table}

\subsection{Experiment 2: Event Framing Method}
\label{section:experiment_framing_method}
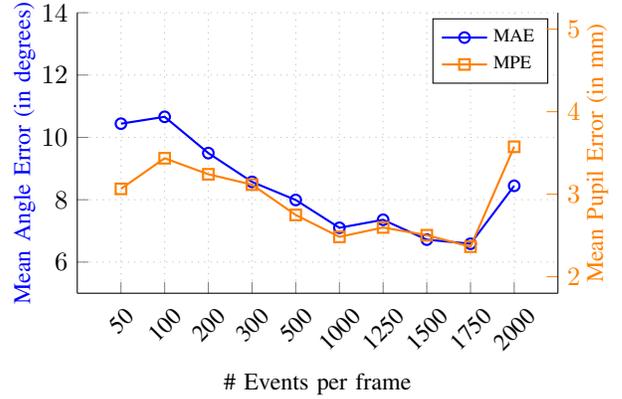
\begin{figure}
\centering
\begin{tikzpicture}
    \begin{axis}[
        width=0.9\linewidth,
        height=0.6\linewidth,
        xlabel={\# Events per frame},
        xlabel style={font=\small},
        ylabel={Mean Angle Error (in degrees)},
        ylabel style={yshift=-1ex, color=blue, font=\small},
        xmin=0, xmax=11,
        ymin=5, ymax=14,
        xtick={1,2,3,4,5,6,7,8,9,10},
        xticklabels={50, 100, 200, 300, 500, 1000, 1250, 1500, 1750, 2000},
        xticklabel style={font=\small, rotate=45},
        legend style={font=\scriptsize},
        grid=both,
        grid style={dotted},
        every axis plot/.append style={thick},
        legend cell align={left},
        axis y line*=left,
        axis x line*=bottom
    ]

    \addplot[
        mark=o,
        color=blue,
        solid,
    ] coordinates {
        (1, 10.44)
        (2, 10.66)
        (3, 9.496)
        (4, 8.572)
        (5, 7.992)
        (6, 7.100)
        (7, 7.357)
        (8, 6.719)
        (9, 6.588)
        (10, 8.446)
    };
    \label{plt:mae_event_count}
    \addlegendentry{MAE};
    
    \addlegendimage{/pgfplots/refstyle=plt:mpe_event_count}\addlegendentry{MPE};
    
    \end{axis}

    \begin{axis}[
        width=0.9\linewidth,
        height=0.6\linewidth,
        xlabel={\# Events per frame},
        xlabel style={font=\small},
        ylabel={Mean Pupil Error (in mm)},
        ylabel style={yshift=1ex, color=orange, font=\small},
        xmin=0, xmax=11,
        ymin=1.8, ymax=5.2,
        xtick={1,2,3,4,5,6,7,8,9,10},
        xticklabels={50, 100, 200, 300, 500, 1000, 1250, 1500, 1750, 2000},
        xticklabel style={font=\small, rotate=45},
        axis y line*=right,
        axis x line=none,
        ytick style={color=orange},
        yticklabel style={color=orange},
        every axis plot/.append style={thick},
    ]

    \addplot[
        mark=square,
        color=orange,
        solid,
    ] coordinates {
        (1, 3.065)
        (2, 3.436)
        (3, 3.242)
        (4, 3.116)
        (5, 2.749)
        (6, 2.483)
        (7, 2.597)
        (8, 2.503)
        (9, 2.364)
        (10, 3.575)
    };
    \label{plt:mpe_event_count}
    \end{axis}
\end{tikzpicture}
\caption{Mean Angle Error (MAE) and Mean Pupil Error (MPE) vs. the number of events per aggregated frame in the input sequence}
\label{fig:mae_mpe_event_count}
\end{figure}

We assess the effects of converting event streams into frames using two methods: fixed event counts per frame and fixed time windows per frame.
Note that the framing method and granularity affect the total number of frames in the dataset. So, the test and validation sets need to be redrawn for each framing method, meaning the results from this experiment all use a different test subset.
Table~\ref{tab:ablation_study_framing} and Figure~\ref{fig:mae_mpe_event_count} present results, indicating that framing method and granularity significantly influence accuracy, with a fixed event count of 1750 achieving the best accuracy ($MAE=6.588\degree, MPE=2.364\text{mm}$). Fixed time windows perform worse overall, with the optimal accuracy ($MAE=8.242\degree, MPE=3.093\text{mm}$) achieved at 10ms, exactly matching the update rate for the gaze references from the Tobii Pro Glasses 3 at $100\text{Hz}$, meaning the least interpolation.

\begin{table}[!t]
\caption{Ablation Study Results of Varying Framing Method for GazeSCRNN}
\label{tab:ablation_study_framing}
\centering
\scriptsize
\begin{tabular}{c|c||c|c|c}
\hline
\textbf{Method} & \textbf{\# Events / Length} & \textbf{MAE} & \textbf{MPE} & \textbf{MFR} \\
\hline
\hline
Event count & 50 events & $10.44\degree$ & $3.065\text{mm}$ & $0.01748$ \\
\hline
Event count & 100 events & $10.66\degree$ & $3.436\text{mm}$ & $0.04163$ \\
\hline
Event count & 200 events & $9.496\degree$ & $3.242\text{mm}$ & $0.02253$ \\
\hline
Event count & 300 events & $8.572\degree$ & $3.116\text{mm}$ & $0.02244$ \\
\hline
Event count & 500 events & $7.992\degree$ & $2.749\text{mm}$ & $0.02369$ \\
\hline
Event count & 1000 events & $7.100\degree$ & $2.483\text{mm}$ & $0.04289$ \\
\hline
Event count & 1250 events & $7.357\degree$ & $2.597\text{mm}$ & $0.03287$ \\
\hline
Event count & 1500 events & $6.719\degree$ & $2.503\text{mm}$ & $0.03451$ \\
\hline
Event count & 1750 events & $\mathbf{6.588\degree}$ & $2.364\text{mm}$ & $0.04504$ \\
\hline
Event count & 2000 events & $8.446\degree$ & $3.575\text{mm}$ & $0.03579$ \\
\hline
Time window & 1ms & $11.03\degree$ & $3.828\text{mm}$ & $0.01348$ \\
\hline
Time window & 5ms & $9.575\degree$ & $3.934\text{mm}$ & $0.01849$ \\
\hline
Time window & 10ms & $8.242\degree$ & $3.093\text{mm}$ & $0.01855$ \\
\hline
Time window & 20ms & $13.33\degree$ & $4.805\text{mm}$ & $0.02532$ \\
\hline
\end{tabular}
\end{table}

\subsection{Experiment 3: Neuron Types}
\label{section:experiment_neuron_types}
This experiment investigates the impact of neuron complexity on model performance by comparing Adaptive Leaky-Integrate-and-Fire (ALIF) neurons, as described in Section~\ref{section:spiking_neurons}, to the simpler Parametric Leaky-Integrate-and-Fire (PLIF) neurons. For this, all ALIF layers in the GazeSCRNN architecture were replaced with PLIF layers, resulting in a model with only $790\text{k}$ parameters, compared to the $2.3\text{M}$ parameters in the original architecture. Both configurations were evaluated using aggregated frames with 300 and 1750 events, chosen based on findings in Section~\ref{section:experiment_framing_method}, and were trained using either FPTT with 8 truncated BPTT time steps, or 12 truncated BPTT time steps without FPTT, based on findings in Section~\ref{section:experiment_fptt}.

The results, shown in Table~\ref{tab:ablation_study_plif}, indicate that models using PLIF neurons are significantly less accurate than those with ALIF neurons in all tested configurations. While the reduction in parameters highlights the computational efficiency of the PLIF-based model, this simplification leads to significant increases in MAE and MPE. These findings underscore the importance of the adaptive mechanisms provided by ALIF neurons, which appear to capture temporal dependencies more effectively.

\begin{table}[!t]
\caption{Ablation Study Results of Using Different Spiking Neurons for GazeSCRNN}
\label{tab:ablation_study_plif}
\centering
\scriptsize
\begin{tabular}{c|c|c||c|c|c}
\hline
\textbf{Neurons} & \textbf{\# Events} & \textbf{Training} & \textbf{MAE} & \textbf{MPE} & \textbf{MFR} \\
\hline
\hline
ALIF & 300 & FPTT, $T=8$ & $8.572\degree$ & $3.116\text{mm}$ & $0.02244$ \\
\hline
PLIF & 300 & FPTT, $T=8$ & $12.89\degree$ & $4.218\text{mm}$ & $0.02919$ \\
\hline
ALIF & 300 & $T=12$ & $8.463\degree$ & $3.064\text{mm}$ & $0.03457$ \\
\hline
PLIF & 300 & $T=12$ & $12.14\degree$ & $3.759\text{mm}$ & $0.06570$ \\
\hline
ALIF & 1750 & FPTT, $T=8$ & $\mathbf{6.588\degree}$ & $2.364\text{mm}$ & $0.04504$ \\
\hline
PLIF & 1750 & FPTT, $T=8$ & $8.451\degree$ & $3.252\text{mm}$ & $0.04007$ \\
\hline
ALIF & 1750 & $T=12$ & $6.965\degree$ & $2.740\text{mm}$ & $0.03508$ \\
\hline
PLIF & 1750 & $T=12$ & $10.94\degree$ & $4.081\text{mm}$ & $0.04404$ \\
\hline
\end{tabular}
\end{table}

\subsection{Experiment 4: Excluding Unreliable Interpolated References}
\label{section:experiment_unreliable_references}
This experiment examines the exclusion of unreliable interpolated gaze references from loss and metric computations. Unreliable references arise when long intervals between ground truth gaze vectors from the Tobii Pro Glasses 3 require extensive interpolation.
We define thresholds for acceptable interpolation durations, masking references exceeding these thresholds while retaining their sequence data for membrane potential and recurrent connection dynamics. Results, summarized in Table~\ref{tab:ablation_study_references_time_window}, demonstrate that excluding unreliable references can improve MAE and MPE, particularly when paired with larger event counts per frame and tighter thresholds. However, this is not necessarily the case, as can be seen in the result with $300$ events per frame and a maximum threshold of $20\text{ms}$, which has worse accuracy.

\begin{table}[!t]
\caption{Ablation Study Results of Excluding Unreliable Interpolated References from Loss and Metrics for GazeSCRNN}
\label{tab:ablation_study_references_time_window}
\centering
\scriptsize
\begin{tabular}{c|c|c||c|c|c}
\hline
\textbf{Threshold} & \textbf{\# Events} & \textbf{Included} & \textbf{MAE} & \textbf{MPE} & \textbf{MFR} \\
\hline
\hline
N/A & 300 events & $100\%$ & $8.572\degree$ & $3.116\text{mm}$ & $0.02244$ \\
\hline
N/A & 1500 events & $100\%$ & $6.719\degree$ & $2.503\text{mm}$ & $0.03451$ \\
\hline
N/A & 1750 events & $100\%$ & $6.588\degree$ & $2.364\text{mm}$ & $0.04504$ \\
\hline
50ms & 300 events & $84.98\%$ & $8.396\degree$ & $2.829\text{mm}$ & $0.02337$ \\
\hline
50ms & 1500 events & $84.98\%$ & $6.156\degree$ & $2.199\text{mm}$ & $0.03263$ \\
\hline
50ms & 1750 events & $84.99\%$ & $6.56\degree$ & $2.482\text{mm}$ & $0.05066$ \\
\hline
20ms & 300 events & $59.87\%$ & $8.807\degree$ & $2.650\text{mm}$ & $0.02498$ \\
\hline
20ms & 1500 events & $59.88\%$ & $\mathbf{6.034\degree}$ & $2.094\text{mm}$ & $0.03391$ \\
\hline
20ms & 1750 events & $59.89\%$ & $6.216\degree$ & $2.369\text{mm}$ & $0.07050$ \\
\hline
\end{tabular}
\end{table}

\subsection{Experiment 5: Using Full Size Inputs}
\label{section:experiment_full_size}
In the default setup, all models are trained using input frames downscaled by a factor of two, resulting in dimensions of $173 \times 130 \times 2$. For this experiment, we evaluate the impact of training the model on full-size frames with dimensions of $346 \times 260 \times 2$, containing either $300$ or $1750$ events per frame.
To accommodate the increased input size, an additional convolutional block is introduced to extract spatial features at the larger scale, followed by a max-pooling layer to downscale the spatial dimensions by a factor of two.
The results, shown in Table~\ref{tab:ablation_study_full_size_inputs}, indicate that using full-size inputs does not significantly improve accuracy for an event count of 300 events per frame. Furthermore, for an event count of 1750 events per frame, using full-size inputs significantly decreases accuracy.

\begin{table}[!t]
\caption{Ablation Study Results of Using Full Size Inputs for GazeSCRNN}
\label{tab:ablation_study_full_size_inputs}
\centering
\scriptsize
\begin{tabular}{c|c||c|c|c}
\hline
\textbf{Full Size Input} & \textbf{\# Events} & \textbf{MAE} & \textbf{MPE} & \textbf{MFR} \\
\hline
\hline
\xmark & 300 events & $8.572\degree$ & $3.116\text{mm}$ & $0.02244$ \\
\hline
\cmark & 300 events & $8.396\degree$ & $2.676\text{mm}$ & $0.01083$ \\
\hline
\xmark & 1750 events & $\mathbf{6.588\degree}$ & $2.364\text{mm}$ & $0.04504$ \\
\hline
\cmark & 1750 events & $8.355\degree$ & $3.351\text{mm}$ & $0.07456$ \\
\hline
\end{tabular}
\end{table}

\section{Discussion}
\label{section:discussion}
The GazeSCRNN model highlights the potential of spiking neural networks (SNNs) for event-based gaze tracking, but also reveals several limitations that suggest areas for improvement. A significant challenge lies in the reliance on gaze reference data generated by the Tobii Pro Glasses 3 gaze tracker. We observed prolonged periods where it failed to provide valid gaze vectors, leading to interpolation between widely separated ground truth values, giving unreliable targets. Excluding unreliable interpolated references above a time window threshold improves model accuracy but reduces the amount of usable training data, particularly when tight thresholds are applied. This trade-off underscores a broader limitation in the quality of ground truth data and suggests a need for alternative strategies for generating robust ground truth targets.

The preprocessing method for framing event streams also presents important trade-offs. Two framing methods were examined in this paper: fixed event counts and fixed time windows. Fixed event count framing results in a dynamic frame rate that adapts to the rate of incoming events from the Dynamic Vision Sensor (DVS) camera, aligning with the asynchronous nature of the DVS camera, by only providing updates when there are observed eye movements. This allows the model to focus on meaningful changes in the user's gaze, rather than processing redundant information. However, this approach means that each frame represents a variable timescale, which can complicate the interpretation of temporal information. Despite this, fixed event count framing demonstrated superior accuracy compared to fixed time window framing. This is likely because fixed event counts ensure that every frame contains sufficient information to make accurate predictions, while fixed time windows may yield frames with highly variable and often insufficient event counts. 

Another consideration is the relationship between framing choices and the gaze tracker's update rate. The number of events per frame or the duration of the time windows determines the frame rate of the input sequence, which in turn can affect the output update rate of the gaze tracker. For example, if the system requires the aggregation of frames before making predictions, the gaze tracker's update rate becomes tied to the input frame rate, potentially introducing latency. Future research could explore alternative methods, such as rolling windows or asynchronous spike processing, to address these challenges and provide more frequent predictions.

The computational demands of the GazeSCRNN model represent another area for improvement. Although the architecture is relatively simple, it relies on the complex and trainable spiking behaviour of the Adaptive Leaky-Integrate-and-Fire (ALIF) neurons to provide versatility and accuracy. While ALIF neurons are integral to capturing temporal dependencies effectively, they significantly increase the parameter count compared to simpler Parametric Leaky-Integrate-and-Fire (PLIF) neurons. This added complexity poses challenges for deploying the model on resource-constrained hardware. Future work could investigate hybrid neuron architectures or alternative adaptive mechanisms that maintain the temporal modelling capabilities of ALIF neurons while reducing computational overhead. Moreover, emerging architectures such as spiking transformers~\cite{zhouSpikformerWhenSpiking2022,yaoSpikedrivenTransformer2023} and spiking state space models~\cite{liSpikeMbaMultiModalSpiking2024} could be explored for event-based gaze tracking, potentially offering novel ways to improve both accuracy and efficiency.

To unleash the full potential of the speed and energy efficiency of event-based gaze tracking, the GazeSCRNN model would ideally be implemented on neuromorphic hardware optimized for spiking neural networks. Such hardware could leverage the sparsity of spiking computations, enhancing both energy efficiency and processing speed~\cite{pfeifferDeepLearningSpiking2018}. We recommend this as a focus for further research, evaluating the performance of GazeSCRNN in terms of latency and energy consumption, and benchmarking it against existing gaze tracking systems. This would provide critical insights into the practical applicability of our work in real-world settings.

\section{Conclusion}
\label{section:conclusion}
This paper introduces GazeSCRNN, a spiking convolutional recurrent neural network designed for event-based gaze tracking, leveraging the high temporal resolution and asynchronous nature of Dynamic Vision Sensor (DVS) cameras.

This work demonstrates the viability and potential of SNNs for event-based gaze tracking, while identifying key challenges and opportunities for improvement. By addressing these limitations, the GazeSCRNN model can contribute to advancing the field of gaze tracking, particularly for applications requiring high temporal precision and energy efficiency.
The findings of this work pave the way for future research endeavours, such as enhancing the quality of ground truth data, exploring alternative framing strategies such as rolling windows or asynchronous spike processing, and implementing the model on neuromorphic hardware.

\bibliographystyle{IEEEtran}
\bibliography{IEEEabrv,references}

\begin{thebibliography}{10}
\providecommand{\url}[1]{#1}
\csname url@samestyle\endcsname
\providecommand{\newblock}{\relax}
\providecommand{\bibinfo}[2]{#2}
\providecommand{\BIBentrySTDinterwordspacing}{\spaceskip=0pt\relax}
\providecommand{\BIBentryALTinterwordstretchfactor}{4}
\providecommand{\BIBentryALTinterwordspacing}{\spaceskip=\fontdimen2\font plus
\BIBentryALTinterwordstretchfactor\fontdimen3\font minus
  \fontdimen4\font\relax}
\providecommand{\BIBforeignlanguage}[2]{{%
\expandafter\ifx\csname l@#1\endcsname\relax
\typeout{** WARNING: IEEEtran.bst: No hyphenation pattern has been}%
\typeout{** loaded for the language `#1'. Using the pattern for}%
\typeout{** the default language instead.}%
\else
\language=\csname l@#1\endcsname
\fi
#2}}
\providecommand{\BIBdecl}{\relax}
\BIBdecl

\bibitem{hansenEyeBeholderSurvey2010}
D.~W. Hansen and Q.~Ji, ``In the {{Eye}} of the {{Beholder}}: {{A Survey}} of
  {{Models}} for {{Eyes}} and {{Gaze}},'' \emph{IEEE Transactions on Pattern
  Analysis and Machine Intelligence}, vol.~32, no.~3, pp. 478--500, Mar. 2010.

\bibitem{liuEyeBeholderSurvey2022}
J.~Liu, J.~Chi, H.~Yang, and X.~Yin, ``In the eye of the beholder: {{A}} survey
  of gaze tracking techniques,'' \emph{Pattern Recognition}, vol. 132, p.
  108944, Dec. 2022.

\bibitem{duchowskiEyeTrackingMethodology2017}
A.~T. Duchowski, \emph{Eye {{Tracking Methodology}}}.\hskip 1em plus 0.5em
  minus 0.4em\relax Cham: Springer International Publishing, 2017.

\bibitem{gallegoEventBasedVisionSurvey2022}
G.~Gallego, T.~Delbr{\"u}ck, G.~Orchard, C.~Bartolozzi, B.~Taba, A.~Censi,
  S.~Leutenegger, A.~J. Davison, J.~Conradt, K.~Daniilidis, and D.~Scaramuzza,
  ``Event-{{Based Vision}}: {{A Survey}},'' \emph{IEEE Transactions on Pattern
  Analysis and Machine Intelligence}, vol.~44, no.~1, pp. 154--180, Jan. 2022.

\bibitem{lichtsteiner128x128120DB2008}
P.~Lichtsteiner, C.~Posch, and T.~Delbruck, ``A 128x128 120 {{dB}} 15 {$M$}s
  {{Latency Asynchronous Temporal Contrast Vision Sensor}},'' \emph{IEEE
  Journal of Solid-State Circuits}, vol.~43, no.~2, pp. 566--576, Feb. 2008.

\bibitem{shahsavariAdvancementsSpikingNeural2023}
M.~Shahsavari, D.~Thomas, M.~{van Gerven}, A.~Brown, and W.~Luk, ``Advancements
  in spiking neural network communication and synchronization techniques for
  event-driven neuromorphic systems,'' \emph{Array}, vol.~20, p. 100323, Dec.
  2023.

\bibitem{pfeifferDeepLearningSpiking2018}
M.~Pfeiffer and T.~Pfeil, ``Deep {{Learning With Spiking Neurons}}:
  {{Opportunities}} and {{Challenges}},'' \emph{Frontiers in Neuroscience},
  vol.~12, Oct. 2018.

\bibitem{zhaoEVEyeRethinkingHighfrequency2023}
G.~Zhao, Y.~Yang, J.~Liu, N.~Chen, Y.~Shen, H.~Wen, and G.~Lan, ``{{EV-Eye}} :
  {{Rethinking}} high-frequency eye tracking through the lenses of event
  cameras,'' in \emph{Thirty-Seventh {{Conference}} on {{Neural Information
  Processing Systems}} ({{NeurIPS}}), 2023}, New Orleans, USA, Sep. 2023.

\bibitem{neftciSurrogateGradientLearning2019}
E.~O. Neftci, H.~Mostafa, and F.~Zenke, ``Surrogate {{Gradient Learning}} in
  {{Spiking Neural Networks}}: {{Bringing}} the {{Power}} of {{Gradient-Based
  Optimization}} to {{Spiking Neural Networks}},'' \emph{IEEE Signal Processing
  Magazine}, vol.~36, no.~6, pp. 51--63, Nov. 2019.

\bibitem{bellecLongShorttermMemory2018}
G.~Bellec, D.~Salaj, A.~Subramoney, R.~Legenstein, and W.~Maass, ``Long
  short-term memory and {{Learning-to-learn}} in networks of spiking neurons,''
  in \emph{Advances in {{Neural Information Processing Systems}}},
  vol.~31.\hskip 1em plus 0.5em minus 0.4em\relax Curran Associates, Inc.,
  2018.

\bibitem{yinAccurateOnlineTraining2022}
B.~Yin, F.~Corradi, and S.~M. Bohte, ``Accurate online training of dynamical
  spiking neural networks through {{Forward Propagation Through Time}},'' Nov.
  2022.

\bibitem{jiangAppearanceBasedGazeTracking2019}
J.~Jiang, X.~Zhou, S.~Chan, and S.~Chen, ``Appearance-{{Based Gaze Tracking}}:
  {{A Brief Review}},'' in \emph{Intelligent {{Robotics}} and
  {{Applications}}}, H.~Yu, J.~Liu, L.~Liu, Z.~Ju, Y.~Liu, and D.~Zhou,
  Eds.\hskip 1em plus 0.5em minus 0.4em\relax Cham: Springer International
  Publishing, 2019, pp. 629--640.

\bibitem{angelopoulosEventBasedNearEyeGaze2021}
A.~N. Angelopoulos, J.~N. Martel, A.~P. Kohli, J.~Conradt, and G.~Wetzstein,
  ``Event-{{Based Near-Eye Gaze Tracking Beyond}} 10,000 {{Hz}},'' \emph{IEEE
  Transactions on Visualization and Computer Graphics}, vol.~27, no.~5, pp.
  2577--2586, May 2021.

\bibitem{ryanRealtimeFaceEye2021}
C.~Ryan, B.~O'Sullivan, A.~Elrasad, A.~Cahill, J.~Lemley, P.~Kielty, C.~Posch,
  and E.~Perot, ``Real-time face \& eye tracking and blink detection using
  event cameras,'' \emph{Neural Networks}, vol. 141, pp. 87--97, Sep. 2021.

\bibitem{chen3ETEfficientEventbased2023}
Q.~Chen, Z.~Wang, S.-C. Liu, and C.~Gao, ``{{3ET}}: {{Efficient Event-based Eye
  Tracking}} using a {{Change-Based ConvLSTM Network}},'' in \emph{2023 {{IEEE
  Biomedical Circuits}} and {{Systems Conference}} ({{BioCAS}})}, Oct. 2023,
  pp. 1--5.

\bibitem{jiangEyeTrackingBased2024a}
Y.~Jiang, W.~Wang, L.~Yu, and C.~He, ``Eye {{Tracking Based}} on {{Event
  Camera}} and {{Spiking Neural Network}},'' \emph{Electronics}, vol.~13,
  no.~14, p. 2879, Jan. 2024.

\bibitem{bonazziRetinaLowPowerEye2024}
P.~Bonazzi, S.~Bian, G.~Lippolis, Y.~Li, S.~Sheik, and M.~Magno, ``Retina :
  {{Low-Power Eye Tracking}} with {{Event Camera}} and {{Spiking Hardware}},''
  in \emph{Proceedings of the {{IEEE}}/{{CVF Conference}} on {{Computer
  Vision}} and {{Pattern Recognition}}}, 2024, pp. 5684--5692.

\bibitem{inivationagDAVIS346Documentation2024}
{iniVation AG}, ``{{DAVIS346 Documentation}},'' Nov. 2024.

\bibitem{tobiiabTobiiProGlasses2024}
{Tobii AB}, ``Tobii {{Pro Glasses}} 3 {{Product Description}},'' Jun. 2024.

\bibitem{yinEffectiveEfficientComputation2020}
B.~Yin, F.~Corradi, and S.~M. Boht{\'e}, ``Effective and {{Efficient
  Computation}} with {{Multiple-timescale Spiking Recurrent Neural
  Networks}},'' in \emph{International {{Conference}} on {{Neuromorphic
  Systems}} 2020}, ser. {{ICONS}} 2020.\hskip 1em plus 0.5em minus 0.4em\relax
  New York, NY, USA: Association for Computing Machinery, Jul. 2020, pp. 1--8.

\bibitem{kagTrainingRecurrentNeural2021}
A.~Kag and V.~Saligrama, ``Training {{Recurrent Neural Networks}} via {{Forward
  Propagation Through Time}},'' in \emph{Proceedings of the 38th
  {{International Conference}} on {{Machine Learning}}}.\hskip 1em plus 0.5em
  minus 0.4em\relax PMLR, Jul. 2021, pp. 5189--5200.

\bibitem{varposhtiEnergyEfficientSpikingRecurrent2024}
M.~H. Varposhti, M.~Shahsavari, and M.~van Gerven, ``Energy-{{Efficient Spiking
  Recurrent Neural Network}} for {{Gesture Recognition}} on {{Embedded
  GPUs}},'' Aug. 2024.

\bibitem{zhouSpikformerWhenSpiking2022}
Z.~Zhou, Y.~Zhu, C.~He, Y.~Wang, S.~Yan, Y.~Tian, and L.~Yuan, ``Spikformer:
  {{When Spiking Neural Network Meets Transformer}},'' in \emph{The {{Eleventh
  International Conference}} on {{Learning Representations}}}, Sep. 2022.

\bibitem{yaoSpikedrivenTransformer2023}
M.~Yao, J.~Hu, Z.~Zhou, L.~Yuan, Y.~Tian, B.~Xu, and G.~Li, ``Spike-driven
  {{Transformer}},'' \emph{Advances in Neural Information Processing Systems},
  vol.~36, pp. 64\,043--64\,058, Dec. 2023.

\bibitem{liSpikeMbaMultiModalSpiking2024}
W.~Li, X.~Hong, R.~Xiong, and X.~Fan, ``{{SpikeMba}}: {{Multi-Modal Spiking
  Saliency Mamba}} for {{Temporal Video Grounding}},'' May 2024.

\end{thebibliography}

\end{document}